\documentclass[10pt,twocolumn,letterpaper]{article}

\usepackage[pagenumbers]{cvpr} 

\usepackage[dvipsnames]{xcolor}

\usepackage{arydshln}

\definecolor{cvprblue}{rgb}{0.21,0.49,0.74}
\usepackage[pagebackref,breaklinks,colorlinks,citecolor=cvprblue]{hyperref}

\def\paperID{9710} 
\def\confName{CVPR}
\def\confYear{2024}

\def \ie{{\em i.e.}}
\def \eg{{\em e.g.}}
\def \etal{{\em et al.}}

\usepackage{arydshln} 
\usepackage{booktabs}
\usepackage{multirow}

\usepackage{amsthm}
\usepackage{mathtools}
\usepackage{algorithm}
\usepackage{bbding}
\usepackage{listings}
\usepackage{subcaption}
\usepackage{overpic}
\input{def.set}
\usepackage{pifont}
\usepackage{color}
\usepackage{amssymb}
\usepackage{graphicx}
\usepackage{wrapfig}
\usepackage{makecell}
\usepackage{colortbl}
\usepackage{multirow}
\usepackage{fixltx2e}
\usepackage{subcaption}
\usepackage{amsmath}
\usepackage{etoolbox}
\usepackage{multicol}
\usepackage{refcount}
\usepackage{enumitem}
\usepackage{hyperref}
\usepackage{colortbl}\definecolor{mygray}{gray}{.9}

\usepackage{color}

\usepackage{listings}
\usepackage{newfloat}
\definecolor{codegreen}{rgb}{0.5,0.6,0.6}
\definecolor{codegray}{rgb}{0.5,0.5,0.5}
\definecolor{codepurple}{rgb}{0.58,0,0.82}
\definecolor{backcolour}{rgb}{0.95,0.95,0.92}

\usepackage{algorithm}
\usepackage{algorithmic}
\lstdefinestyle{mystyle}{
  commentstyle=\color{codegreen},
  keywordstyle=\color{magenta},
  numberstyle=\tiny\color{codegray},
  stringstyle=\color{codepurple},
  basicstyle=\linespread{0.8}\ttfamily\scriptsize,
  breakatwhitespace=false,         
  breaklines=true,                 
  captionpos=b,                    
  keepspaces=false,  
  showspaces=false,                
  showstringspaces=false,
  showtabs=false,                  
  tabsize=1
}
\newsavebox{\mycode}

\usepackage{etoolbox}
\makeatletter
\AfterEndEnvironment{algorithm}{\let\@algcomment\relax}
\AtEndEnvironment{algorithm}{\kern2pt\hrule\relax\vskip3pt\@algcomment}
\let\@algcomment\relax
\newcommand\algcomment[1]{\def\@algcomment{\footnotesize#1}}
\renewcommand\fs@ruled{\def\@fs@cfont{\bfseries}\let\@fs@capt\floatc@ruled
  \def\@fs@pre{\hrule height.8pt depth0pt \kern2pt}%
  \def\@fs@post{}%
  \def\@fs@mid{\kern2pt\hrule\kern2pt}%
  \let\@fs@iftopcapt\iftrue}
\makeatother

\title{SD-DiT: Unleashing the Power of Self-supervised Discrimination in\\ Diffusion Transformer\thanks{This work was performed at HiDream.ai.}}

\author{Rui Zhu$^{1}$,~~ Yingwei Pan$^{2}$,~~Yehao Li$^2$,~~Ting Yao$^2$,~~Zhenglong Sun$^1$,~~Tao Mei$^2$,~ Chang Wen Chen$^3$  \\
	{\small\centering$^1$ The Chinese University of HongKong, Shenzhen}~~~
	{\small\centering$^2
$ HiDream.ai Inc.}~~~
	{\small\centering$^3$  The Hong Kong Polytechnic University}
	\\
	{\tt\scriptsize ruizhu@link.cuhk.edu.cn, \{pandy, liyehao, tiyao\}@hidream.ai, sunzhenglong@cuhk.edu.cn} \\ {\tt\scriptsize tmei@hidream.ai, changwen.chen@polyu.edu.hk}
	}

\begin{document}
\maketitle
\begin{abstract}

Diffusion Transformer (DiT) has emerged as the new trend of generative diffusion models on image generation. In view of extremely slow convergence in typical DiT, recent breakthroughs have been driven by mask strategy that significantly improves the training efficiency of DiT with additional intra-image contextual learning. Despite this progress, mask strategy still suffers from two inherent limitations: (a) training-inference discrepancy and (b) fuzzy relations between mask reconstruction \& generative diffusion process, resulting in sub-optimal training of DiT. In this work, we address these limitations by novelly unleashing the self-supervised discrimination knowledge to boost DiT training. Technically, we frame our DiT in a teacher-student manner. The teacher-student discriminative pairs are built on the diffusion noises along the same Probability Flow Ordinary Differential Equation (PF-ODE). Instead of applying mask reconstruction loss over both DiT encoder and decoder, we decouple DiT encoder and decoder to separately tackle discriminative and generative objectives. 
In particular, by encoding discriminative pairs with student and teacher DiT encoders, a new discriminative loss is designed to encourage the inter-image alignment in the self-supervised embedding space. After that, student samples are fed into student DiT decoder to perform the typical generative diffusion task.
Extensive experiments are conducted on ImageNet dataset, and our method achieves a competitive balance between training cost and generative capacity.

\end{abstract}    
\section{Introduction} \label{sec:intro}

Recent computer vision field has witnessed the rise of diffusion models~\cite{sohl2015deep,ho2020denoising,song2020score,luo2023semantic} in powerful and scalable generative architectures for image generation. 
Such practical generative model pushes the limits of a series of CV applications, including text-to-image  synthesis~\cite{balaji2022ediffi,saharia2022photorealistic,ramesh2022hierarchical,rombach2022high}, video generation~\cite{gupta2023photorealistic,videoworldsimulators2024,yu2023magvit,ho2022imagen}, and 3D generation \cite{chen2023control3d,yang20233dstyle}.

\begin{figure}[t]
  \footnotesize
  \vspace{-1.5em}
  \centering
  \setlength{\tabcolsep}{0.2mm}
  \renewcommand{\arraystretch}{0.6}
  \subfloat[SD-DiT]{\footnotesize 
  \centering
  \includegraphics[width=0.45\linewidth]{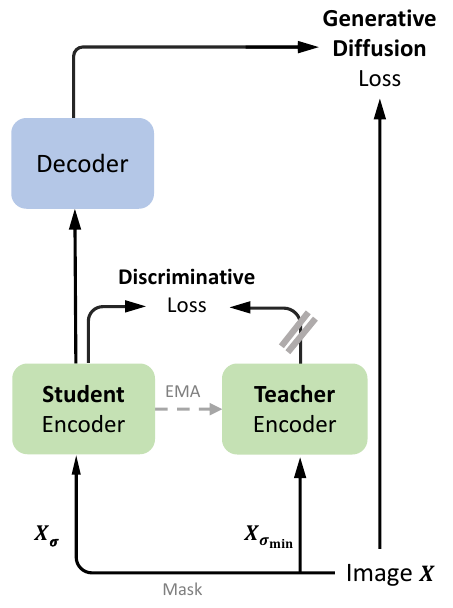}
  \label{fig:did}  
  } 
  \hfill
  \subfloat[MaskDiT]{\footnotesize
  \includegraphics[width=0.51\linewidth]{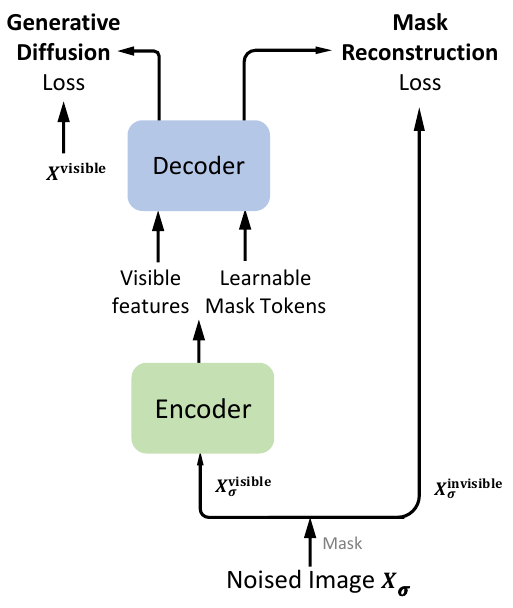}
  \label{fig:maskdit}
  }
    \label{fig:conceptual}
  \vspace{-1em}
  \caption{Conceptual comparison between (a) our SD-DiT and (b) MaskDiT. 
  MaskDiT integrates generative diffusion process with mask reconstruction auxiliary task, and the whole DiT encoder plus decoder are jointly optimized for the two tasks.
  In contrast, our SD-DiT frames mask modeling on the basis of discrimination knowledge distilling in a self-supervised manner, pursuing the inter-image alignment in the joint embedding space of teacher and student encoder. DiT encoder and decoder are decoupled to separately tackle discriminative and generative diffusion objectives.
  }
  \vspace{-2em}
\end{figure}

A recent pioneering practice is the Diffusion Transformer (DiT)~\cite{peebles2023scalable}, which inherits the impressive scaling properties of Transformers~\cite{vaswani2017attention} and significantly improves the capacity \& scalability of diffusion models. Unfortunately, similar to  Vision Transformers~\cite{dosovitskiy2020image}, the training of DiT usually suffers from slow convergence and heavy computation burden issues. The recent works \cite{gao2023masked,zheng2023fast}
then turn their focus on investigating the way to accelerate the training convergence of DiT. Many consider combining the Transformer-based diffusion process with additional mask reconstruction objective via the popular mask strategy~\cite{devlin2018bert,he2022masked,chang2022maskgit}. In particular, MDT~\cite{gao2023masked} simultaneously encodes both the complete and masked image input, in order to enhance the intra-image contextual learning among the associated patches. MaskDiT~\cite{zheng2023fast} integrates generative diffusion process with mask reconstruction auxiliary task to optimize the whole DiT encoder and decoder (see \cref{fig:maskdit}).

Although significantly improved training efficiency is attained, these DiT architectures with mask strategy still struggle with extremely high-fidelity image synthesis and suffer from several inherent limitations. (1) \textbf{Training-inference discrepancy}: Mask strategy inevitably introduces learnable mask tokens for triggering mask reconstruction during DiT training, but no artificial mask token is involved for generative diffusion process at inference. This training-inference discrepancy severely limits the generative capacity of learned DiT. Note that to alleviate such discrepancy, MDT introduces additional dual-path interaction between complete and masked inputs during training, while sacrificing much higher computational and memory cost.
(2) \textbf{Fuzzy relations between mask reconstruction \& generative diffusion process}: Most mask-based DiT structures process both the visible and learnable mask tokens via the same DiT decoder to jointly enable mask reconstruction and generative diffusion process, leaving the inherent different peculiarity of each objective not fully exploited. It is noteworthy that such mask modeling can be regarded as intra-image reconstruction derived from the same data distribution (\eg, from $p_{\sigma\odot\text{mask}}$ to $p_\sigma$ for noised data in MaskDiT). Instead, the generative diffusion process aims to model the translations between the real data distribution $p_\text{data}$ and a different noised data distribution $p_\sigma$.
This issue is also observed in MaskDiT, where mask reconstruction objective will gradually overwhelm generative objective at the late training stage. Accordingly, the joint training of the two distinct objectives with fuzzy relations results in sub-optimal training of DiT when applied to generative task.

To address these limitations, our work paves a new way to frame mask modeling of DiT training on the basis of discrimination knowledge distilling in a self-supervised fashion. We propose a novel Diffusion Transformer model with Self-supervised Discrimination, namely SD-DiT, that pursues highly-efficient learning of DiT with higher generative capacity. Technically, SD-DiT shapes the discrimination knowledge distilling in a teacher-student scheme. As shown in \cref{fig:did}, the input discriminative pairs of teacher and student DiT encoders are derived from different diffusion noises (\ie, $p_{\sigma_1}$ and $p_{\sigma_2}$ along the same Probability Flow Ordinary Differential Equation (PF-ODE) of EDM~\cite{karras2022elucidating}). More importantly, different from typical mask strategy that triggers mask reconstruction objective over both DiT encoder and decoder, SD-DiT decouples DiT encoder and decoder to separately perform discrimination knowledge distilling and generative diffusion process. Our launching point is to fully exploit the mutual but also fuzzy relations between self-supervised discrimination distillation and generative diffusion process through such decoupled DiT design. Eventually, we devise a new discriminative loss to enforce the inter-image alignment of encoded visible tokens between teacher and student DiT encoders in the joint embedding space. Next, SD-DiT only feeds student samples into student DiT decoder for performing the conventional generative diffusion objective. Note that here our discriminative loss can be interpreted as inter-image translation between teacher sample (approximately real data distribution $p_\text{data}$) and student sample (noised data distribution $p_\sigma$), which better aligns with generative diffusion objective than conventional intra-image mask reconstruction objective. As such, the joint optimization of discriminative and generative diffusion objectives strengthens DiT training both effectively and efficiently.

In the meantime, the student branch (student DiT encoder plus decoder) in our decoupled DiT design completely retains the same regular noise in EDM and modules as in the generative modeling at inference. The additional teacher DiT encoder is simply updated as the Exponential Moving Average (EMA) of student DiT encoder in a light-weight fashion, without incurring a heavy computational burden for self-supervised discrimination. In this way, our SD-DiT not only preserves the training efficiency of mask modeling, but also elegantly circumvents the training-inference discrepancy issue.

The main contribution of this work is the proposal of Diffusion Transformer structure that fully unleashes the power of self-supervised discrimination to facilitate DiT training. This also leads to the elegant view of how a training-efficient DiT architecture should be designed for fully exploiting the mutual but also fuzzy relations between mask modeling and generative diffusion process, and how to bridge the training-inference discrepancy tailored to generative task. Through extensive experiments on ImageNet-256$\times$256, we demonstrate that our SD-DiT consistently seeks a better training speed-performance trade-off when compared to state-of-the-art DiT models.

\section{Related Work} \label{sec:relatedwork}
\noindent\textbf{Diffusion Models.}\quad
Denoising diffusion probabilistic models (DDPMs)~\cite{ho2020denoising} greatly
accelerate the development of generative models, especially the tasks of text conditioned image synthesis~\cite{balaji2022ediffi,saharia2022photorealistic,ramesh2022hierarchical,rombach2022high,nichol2021glide}, image editing~\cite{mokady2023null,parmar2023zero,hertz2022prompt,brooks2023instructpix2pix, lugmayr2022repaint,chen2023controlstyle} 
and personalized image generation~\cite{ruiz2023dreambooth,gal2022image}.   
As a score-based model~\cite{song2019generative,song2020improved}, DDPMs introduce a forward process to gradually add Gaussian noise to the data according to Stochastic Differential Equation~\cite{song2020score}, and the iterative denoising procedures are employed to generate high-quality samples.
To tackle such a time-consuming iterative nature of DDPM, fast sampling strategies~\cite{song2020denoising,salimans2022progressive,lu2022dpm,ho2022classifier,karras2022elucidating} and training diffusion in the latent space~\cite{rombach2022high,wang2023binary} are proposed.
Besides, several innovations for improving the network architecture of diffusion models are attained to handle various challenging generation tasks.
Convolutional UNet~\cite{ronneberger2015u} is the de-facto configuration from recent diffusion models~\cite{ho2020denoising} and ADM~\cite{dhariwal2021diffusion} further boosts the generation quality of UNet with scalable model size,
including the adaptive group normalization~\cite{wu2018group}, the attention blocks~\cite{van2016conditional,salimans2017pixelcnn++} and the residual blocks from BigGAN~\cite{brock2018large}. 

\noindent\textbf{Diffusion Transformers.}\quad
Transformers~\cite{vaswani2017attention} provide a new paradigm to connect various domains across language~\cite{devlin2018bert}, vision~\cite{he2022masked,zhou2021ibot,bao2021beit,long2022sifa,LiPAMI22,yao2023dual,yao2024hiri}, and multi-modalities~\cite{radford2021learning,li2023scaling}, with remarkable scaling properties in terms of model size~\cite{kaplan2020scaling} and pre-training efficiency~\cite{he2022masked}.
Recently, some Transformer-based diffusion models~\cite{yang2022your,bao2023all,jabri2022scalable,peebles2023scalable} are proposed to exploit the advantages of Transformer architecture in diffusion models.
For example, GenViT ~\cite{yang2022your} first presents that Vision Transformer (ViT)~\cite{dosovitskiy2020image} has the potential for image generation.
Based on ViT with long skip connections, U-ViT~\cite{bao2023all} is specifically designed for the diffusion model which is characterized by integrating the time, the specific condition, and the noisy image patches as tokens.
DiT~\cite{peebles2023scalable} systematically studies the scaling behaviors of Transformers under the Latent Diffusion Models (LDMs)~\cite{rombach2022high} framework, and achieves better generation quality than the U-Net counterparts with a scaling-up high-capacity backbone.
In this work, we take the conventional DiT blocks as backbone network and the generative diffusion task is implemented as LDMs. 

\noindent\textbf{Self-supervised Learning with Diffusion Models.}\quad
With the dominant status of Transformers in vision and language,
the mask strategy from self-supervised  learning~\cite{devlin2018bert,he2022masked,bao2021beit} has greatly propelled the development of generative models.
Following the paradigm of bidirectional generative modeling~\cite{radford2018improving}, 
MaskGiT~\cite{chang2022maskgit} and MUSE~\cite{chang2023muse} aim at predicting randomly masked visual tokens which were first tokenized from images by a discrete VQ-VAE~\cite{van2017neural,esser2021taming}. Iterative decoding is further utilized to rapidly generate an image. 
Moreover, MAGE~\cite{li2023mage} employs such masked token modeling to unify representation pre-training and image generation. 
On the other hand, diffusion models built upon Transformers~\cite{peebles2023scalable,dosovitskiy2020image} could be well integrated with the mask image modeling~\cite{wei2023diffusion}. For example, inspired by MAE~\cite{he2022masked}, MDT~\cite{gao2023masked} and Mask-DiT~\cite{zheng2023fast} take advantage of the asymmetrical encoder-decoder of MAE and add the learning objective loss of reconstructing masked tokens (without discrete tokenizers) to the original generative diffusion loss.
    Such combination with mask modeling remarkably improves the training efficiency and the contextual reasoning ability of the  Diffusion Transformer (\ie, DiT).
It is noteworthy that mask modeling is built upon the intra-view reconstruction
while the typical self-supervised methods with discriminative joint embedding pretraining~\cite{grill2020bootstrap,he2020momentum,chen2020simple,caron2020unsupervised,caron2021emerging,chen2021empirical} focus on the inter-view alignment (invariance).   
Different from existing mask strategy with intra-image contextual learning, our SD-DiT paves a new way to endow mask modeling in DiT with self-supervised discrimination ability via inter-image alignment.

\newcommand{\x}{{\pmb{x}}}
\newcommand{\w}{\pmb{w}}
\newcommand{\f}{\pmb{f}}
\newcommand{\s}{\pmb{s}}
\newcommand{\n}{\pmb{n}}
\newcommand{\e}{\pmb{e}}
\newcommand{\vv}{\pmb{v}}
\newcommand{\eps}{\pmb{\epsilon}}
\newcommand{\I}{\mathbf{I}}

\section{Approach} \label{sec:method}
In this paper, we devise a Diffusion Transformer with Self-supervised Discrimination (SD-DiT) to frame mask modeling in efficient DiT training as self-supervised discrimination knowledge distilling. This section starts with a brief review of the preliminaries of diffusion models. Then, the overall decoupled architecture for discriminative and generative objectives is elaborated. After that, two different kinds of objectives for generative diffusion process and mask modeling, \ie, generative loss and discriminative loss, are introduced. Finally, the overall objective of SD-DiT at the training stage is provided.

\subsection{Preliminaries}
Diffusion models introduce a forward process to progressively add Gaussian noise to the data distribution $p_\text{data}(\x)$ by a Stochastic Differential Equation (SDE)~\cite{song2020score} over time:
\begin{equation}
\label{f_sde}
    d\x_t = \pmb{\mu}(\x, t)dt + g(t)d\pmb{\omega}_t,
\end{equation}
where $\pmb{\mu}$ and $g$ are the drift and diffusion coefficients, and $\pmb{\omega}$ is the standard Brownian motion.
With the time flowing from 0 to $T$, we denote the marginal distribution of $\x_t$ as $p_t({\x})$.
Based on such an SDE, Song \etal~\cite{song2020score} define the probability flow ordinary differential equation (PF-ODE) in the reverse-time sample generation process: 
\begin{equation}
\label{ode}
    \mathrm{d} {\x_t} = [\pmb{\mu}({\x},t) - \frac{1}{2}g(t)^2 \nabla_{{\x}} \log p_t({\x_t})] \mathrm{d} t.
\end{equation}
Recent EDM~\cite{karras2022elucidating} proposes to add Gaussian
noise with mean zero and standard deviation $\sigma$ into the data distribution.
Specifically, EDM utilizes $p_\sigma({\x})$ instead of $p_t({\x})$
and configures $\pmb{\mu}(\x, t):=\0$ and $g(t):=\sqrt{2t}$ in \cref{ode}.
In this case, the resulting perturbed distribution is given by
$p_\sigma({\x}) = p_\text{data}(\x) \ast \mathcal{N} \big( \mathbf{0}, \sigma^2 \mathbf{I})$, 
where $\ast$ denotes the convolution operation. 
In other words,
the real data $\x_0 \sim p_{\text{data}}(\x)$ can be directly diffused as:
\begin{equation}
\label{addnoise}
\x_\sigma=\x_0 + \n , ~\n \sim \mathcal{N}(\0,\sigma^2\I).
\end{equation}
And the corresponding PF-ODE in EDM is presented as:
\begin{equation}
\label{edm_ode}
    d {\x} = -\sigma \nabla_{{\x}} \log p_\sigma({\x}) d\sigma, ~~~\sigma \in[\sigma_\text{min},\sigma_\text{max}],
\end{equation}
where $\nabla_{{\x}} \log p_\sigma({\x})$ is the score function~\cite{song2020score}. As such, diffusion models are basically regarded as score-based generative models~\cite{song2019generative,song2020denoising,song2020score}.
To avoid numerical instability in ODE solving, $\sigma_\text{min}$ is a small positive value and thus $p_{\sigma_\text{min}}(\x) \approx p_\text{data}(\x)$, while $\sigma_\text{max}$ is large enough so that $p_{\sigma_\text{max}}(\x)$ is close to a tractable Gaussian distribution. 
The training objective of EDM is to minimize the expected $L_2$ denoising loss for $\x_0 \sim p_{\text{data}}(\x)$ separately for each $\sigma$,
by parameterizing a denoiser network as $D_{\theta}$: 
\begin{equation}
\label{edmloss}
    \mathbb{E}_{\x_0 \sim p_{\text{data}}} \mathbb{E}_{\n \sim \mathcal{N}(\0, \sigma^2\I)} \| D_{\theta}(\x_0 + \n, \sigma) - \x_0 \|^2_2.
\end{equation}
The estimated score function is thus measured as:
\begin{equation}
\label{edmscore}
\nabla_{{\x}} \log p_\sigma({\x}) = (D_{\theta}(\x_\sigma, \sigma) -\x_\sigma) / \sigma^2.
\end{equation}
Based on the formulation of EDM, Consistency Models~\cite{song2023consistency,song2023improved} propose
to learn a \emph{consistency function} whose outputs of arbitrary pairs on the PF-ODE trajectory (\cref{edm_ode}) are consistent with $\x_{\sigma_\text{min}}\sim p_{\sigma_\text{min}}(\x) \approx p_\text{data}(\x)$. 
Formally, the \emph{consistency function} is defined as:
\begin{equation}
\label{consistency}
\f: (\x_\sigma, \sigma) \mapsto \x_{\sigma_\text{min}},
\end{equation}
and reflects an important property of \emph{self-consistency}:
\begin{equation}
\label{selfconsistency}
\f({\x_\sigma},\sigma) = \f({\x_{\sigma'}},\sigma'),~~~\sigma,\sigma' \in[\sigma_\text{min},\sigma_\text{max}].
\end{equation}
\begin{figure}
    \centering
    \includegraphics[width=0.8\linewidth]{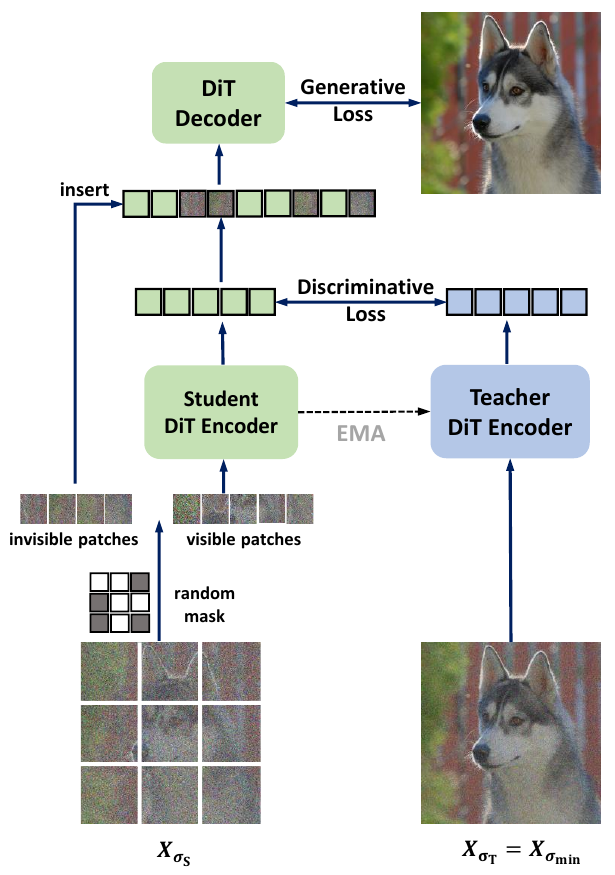}
    \vspace{-1.0em}
    \caption{
    The overview of our SD-DiT. During training, the student view is diffused with regular noise as in EDM formulation, while the teacher view is derived from fixed minimum noise of the consistency function that is close to real data distribution.
    SD-DiT feeds the discriminative pair into teacher and student DiT encoders to perform self-supervised discriminative process within the joint embedding space. Meanwhile, only the student DiT encoder and DiT decoder undertake the generative diffusion process. At inference, all patches are fed into student branch for sampling.}   
    \label{fig:framework}
    \vspace{-1em}
\end{figure}
The diffusion noising schedule in our SD-DiT follows the basic formulation of EDM. 
And the discrimination objective in our SD-DiT is framed on the basis of the theory of the \emph{consistency function}~(\cref{consistency}).
\subsection{Overall Architecture}
The motivation of our SD-DiT is to exploit self-supervised discrimination to facilitate the efficient training of Diffusion Transformer.
\cref{fig:framework} illustrates the overall architecture of SD-DiT, which triggers mask modeling as discrimination knowledge distilling in a teacher-student scheme.
\noindent\textbf{Decoupled Encoder-Decoder Structure.}\quad
Technically, our SD-DiT consists of teacher/student DiT encoders and one DiT decoder, and the core generative objective is framed on the basis of latent space as LDM~\cite{rombach2022high}. The additional discriminative objective is shaped as inter-image alignment among teacher and student DiT encoders in self-supervised joint-embedding space~\cite{caron2021emerging,assran2023self}. Considering the fuzzy relations between mask modeling and generative diffusion process, here we leverage a decoupled encoder-decoder structure to perform the joint training of generative and discriminative objectives, rather than optimizing the whole encoder-decoder with mask reconstruction objective as in existing methods~\cite{zheng2023fast}. Specifically, SD-DiT feeds the discriminative pairs into teacher and student DiT encoders to conduct discrimination knowledge distilling. After that, only student samples are fed into student DiT decoder to perform generative diffusion process. In this decoupled design, the discriminative objective only updates DiT encoder by empowering it with inter-image discriminative capacity. Meanwhile, DiT decoder is solely optimized with generative objective by retaining the same regular noise to nicely mimic the generative diffusion process at inference.

\noindent\textbf{Discriminative Pairs.}\quad
In an effort to trigger discriminative objective, we construct the input discriminative pairs based on the EDM formulation (\cref{addnoise}).
Since the student branch (including student DiT encoder and decoder) will perform both the generative and discriminative objectives, here the student view should be diffused regularly within a large range, similar to MaskDiT~\cite{zheng2023fast}: $\x_{\sigma_\text{S}}=\x_0 + \n, ~\n \sim \mathcal{N}(\0,\sigma_\text{S}^2\I), ~\sigma_\text{S} \in[\sigma_\text{min},\sigma_\text{max}]$.
For the teacher view, we take inspiration from the InfoMin principle~\cite{tian2020makes} in self-supervised learning, and choose the fixed minimum noise of the \emph{consistency function}~\cite{song2023consistency,song2023improved} to construct input samples:
$\x_{\sigma_\text{T}}=\x_0 + \n , ~\n \sim \mathcal{N}(\0,\sigma_\text{min}^2\I)$.
As such, the noised distribution of teacher view can be the closest one to the original data distribution ($\x_{\sigma_\text{T}}\sim p_{\sigma_\text{min}}(\x) \approx p_\text{data}(\x)$) and far away from the noised student view.
Note that we empirically evaluate various teacher noise across $[\sigma_\text{min},\sigma_\text{max}]$ in \cref{sec:ablation}, and attain the similar observations as in InfoMin principle~\cite{tian2020makes}:
The noised teacher view too close to the noised student view could be harmful to self-supervised discriminative learning. Accordingly, we use the fixed minimum  noise for teacher view in practice.

\subsection{Generative Objective}
Inspired by the training efficiency and location contextual awareness~\cite{gao2023masked,zheng2023fast} brought by mask strategy, we follow the typical mask modeling techniques (\eg, \cite{he2022masked}) to frame the generative objective via asymmetric encoder-decoder structure along the student branch. 

\noindent\textbf{Mask Strategy.}\quad
The image will be divided into $n$ non-overlapping patches through the patch embedding layer of DiT.
Let $\mathcal{M}$ denote the binary random mask with the same size of non-overlapping patches.
It is worth noting that MAE and MaskDiT additionally leverage the mask $\mathcal{M}$ to learn additional mask tokens in mask reconstruction auxiliary task. Instead, our SD-DiT solely utilizes the mask $\mathcal{M}$ to separate the noised student view into visible patches $(\vv_{\sigma_\text{S}}=\x_{\sigma_\text{S}}\odot(1-\mathcal{M}))$ and invisible patches $(\pmb{\bar v}_{\sigma_\text{S}} =\x_{\sigma_\text{S}}\odot\mathcal{M})$, where $\odot$ indicates element-wise multiplication on patches. 

\noindent\textbf{Student Branch.}\quad
Given the visible and invisible patches via mask strategy, the student branch applies the typical asymmetric encoder-decoder architecture \cite{he2022masked,zheng2023fast} to improve the training efficiency.
The student DiT encoder can be built with various DiT-Small/Base/XL backbones, while the lightweight student DiT decoder consists of a fixed number of blocks (\ie, 8 DiT blocks, similar to the configurations of MAE~\cite{he2022masked}.).
The student DiT encoder $\mathcal{S}_{\theta}$ only operates over the visible patch and obtains the visible tokens $\mathcal{S}_{\theta}(\vv_{\sigma_\text{S}})$. Then the student decoder $\mathcal{G}_{\theta}$ is fed with the complete token set $\mathcal{H}$. 
Such an asymmetric paradigm with a high mask ratio proposed by MAE~\cite{he2022masked} greatly reduces the training cost because the main computation burden is carried on the large-scale encoder.

\noindent\textbf{Generative Loss.}\quad
Recall that in existing mask modeling techniques (\eg, MAE and MaskDiT), the input token set $\mathcal{H}$ of decoder  commonly augments the visible tokens $\mathcal{S}_{\theta}(\vv_{\sigma_\text{S}})$ with learnable mask tokens, according to the positions of the mask $\mathcal{M}$.
The mask reconstruction auxiliary task is included to recover the learnable mask tokens from the invisible patches $\pmb{\bar v}$.
It is noteworthy that such mask reconstruction objective can benefit the representation learning, but leaves the inherent different peculiarity of mask modeling and generative objectives under-exploited. MaskDiT also points out the fuzzy relations between these two objectives, where mask reconstruction loss will gradually overwhelm the generative objective at the late training stage.

To alleviate this limitation, we discard the mask reconstruction loss and optimize the DiT decoder with only generative loss.
Formally, for the complete token set $\mathcal{H}$, we remove the learnable mask tokens and directly insert the invisible patches $\pmb{\bar v}$ onto the visible tokens $\mathcal{S}_{\theta}(\vv_{\sigma_\text{S}})$, according to the positions of mask $\mathcal{M}$.
Next, the generative loss operates over the compete tokens, which is measured in the form of EDM (\cref{edmloss}):
\begin{equation}
\small
\label{genloss}
    \mathcal{L}_{\text{G}} =\mathbb{E}_{\x_0 \sim p_{\text{data}}} \mathbb{E}_{\n \sim \mathcal{N}(\0, \sigma_\text{S}^2\I)} \| D_{\theta}(\x_0 + \n, \sigma_\text{S}, \mathcal{M}) - \x_0 \|^2_2,
\end{equation}
where $D_{\theta}$ denotes student branch including the student DiT encoder $\mathcal{S}_{\theta}$ and DiT decoder $\mathcal{G}_{\theta}$.

\subsection{Discriminative Objective}
Unlike typical mask modeling with mask reconstruction loss, our SD-DiT paves a new way to frame mask modeling of DiT training on the basis of discrimination knowledge distilling in a self-supervised manner.
Inspired by self-distilling loss in ViT-based self-supervised methods (\ie, DINO~\cite{caron2021emerging} and iBOT~\cite{zhou2021ibot}), we design discriminative loss to enforce the inter-image alignment of encoded visible tokens
between teacher and student DiT encoders.

Specifically, the teacher sample $\x_{\sigma_\text{T}}$ is fed into teacher DiT encoder $\mathcal{T_{\theta'}}$, yielding the output tokens $\mathcal{T_{\theta'}}(\x_{\sigma_\text{T}})$.
Next, SD-DiT performs discriminative loss over the visible tokens between teacher $\e_\text{T}=\mathcal{T_{\theta'}}(\x_{\sigma_\text{T}})$ and student $\e_\text{S}=\mathcal{S}_{\theta}(\vv_{\sigma_\text{S}})$ in the joint encoding space. 
A three-layer projection head $j_\theta$ operates on $\e_\text{S}$ and $\e_\text{T}$ and outputs the softmax probability distribution over $K$ dimensions.
By denoting the distribution on each student and teacher token as $P_{\text{S}_i}$ and $P_{\text{T}_i}$ ($i \in (1-\mathcal{M})$ indicates the index of visible tokens.), the softmax probability distribution of student is measured as:
\begin{equation}
\label{softmax}
P_{\text{S}_i}=\frac{\exp(j_{\theta}(\e_{\text{S}_i})/\tau_\text{S})[k]}{\sum^K_{k=1}\exp(j_{\theta}(\e_{\text{S}_i})/\tau_\text{S})[k]},
\end{equation}
where the student temperature $\tau_\text{S}$ controls the sharpness of the softmax distribution. A similar formulation also holds for teacher: 
$P_{\text{T}_i}$ with teacher temperature $\tau_\text{T}$.
For each visible token $i$, the discrimination loss targets aligning the distribution between teacher and student by minimizing the cross-entropy loss:
\begin{equation}
\label{disloss}
    \mathcal{L}_{\text{D}}(i) = -\sum_k P_{\text{T}_i}\log(P_{\text{S}_i}).
\end{equation}
The final discrimination loss is calculated over all visible patch tokens and the \texttt{[CLS]} token:
\begin{equation}
\small
\label{ourloss}
    \mathcal{L}_{\text{D}} =  \frac{1}{(1-\mathcal{M})}\sum_{i \in (1-\mathcal{M})}\mathcal{L}_{\text{D}}(i) + \mathcal{L}_{\text{D}}(\texttt{[CLS]}).
\end{equation}
Besides, we adopt the centering technique in DINO to avoid feature collapse, where the batch mean statistic is used to whiten the features before softmax during each training iteration. For simplicity, here we leave the details of centering and the complete pseudo-codes to supplementary materials.

In summary, the overall training loss is the combination of discrimination loss and generative loss: $\mathcal{L}_{\text{D}} + \mathcal{L}_{\text{G}}$.
The parameters of student branch (student DiT encoder and decoder) are optimized by this overall loss. And teacher DiT encoder (parameterized as $\mathcal{T_{\theta'}}$) is updated as the exponential moving average (EMA) of student DiT encoder:
$\mathcal{T_{\theta'}} =\beta \mathcal{T_{\theta'}} +(1-\beta) \mathcal{S_{\theta}}.$
Here $\beta$ is a momentum coefficient.  
During training, the teacher is updated by EMA without SGD back-propagation, thereby only requiring extremely lightweight computational cost. At inference, the teacher is completely removed and no burden is introduced.

\section{Experiments} \label{sec:method}

\subsection{Implementation Details}
In this section, we provide the settings of model architecture, training setup, and evaluation details. We list the detailed configurations in supplementary material.

\noindent\textbf{Model Architecture.}\quad
The basic Transformer blocks in our backbone network fully adopt the DiT~\cite{peebles2023scalable} block which fuses conditional time and class embedding with adaptive layer normalization~\cite{perez2018film}. 
We follow the paradigm of LDM~\cite{rombach2022high} and DiT to perform diffusion generation in the latent space of the frozen pre-trained VAE model~\cite{rombach2022high}, which downsamples a $256\times256\times3$ image into a $32\times32\times4$ latent variable.  
Inspired by \cite{he2022masked,zheng2023fast}, we adopt the asymmetric encoder-decoder for generative diffusion process.
The student DiT encoder $\mathcal{S}_{\theta}$ employs DiT-Small/Base/XL-2 (patch size: 2) and the small-scale DiT decoder $\mathcal{G}_{\theta}$ contains 8 DiT blocks, similar to the configurations of MAE.
For the discriminative objective, we mainly follow the settings of iBOT \cite{zhou2021ibot} and DINO~\cite{caron2021emerging}.
The teacher DiT encoder $\mathcal{T_{\theta'}}$ is the EMA of student encoder, and the momentum coefficient increases from 0.996 to 0.999 at the end of training.
The three-layer projection head $j_\theta$ outputs the \texttt{[CLS]} and patch tokens with $K=8,192$ dimension for softmax probability distribution in discriminative loss~\cref{disloss}.

\noindent\textbf{Training Setup.} \quad
Following previous Transformer-based diffusion models~\cite{peebles2023scalable,zheng2023fast,gao2023masked}, we conduct all the experiments on ImageNet-1K with 256$\times$256 resolution and a batch size of 256.
We adopt the most common settings of DiT, \eg, AdamW~\cite{loshchilov2019decoupled} optimizer with a constant $1e-4$ learning rate and no weight decay.
Without specified stating, the mask ratio is set to 0.2 on the student view, and no mask is applied on the teacher view.
No data augmentation is employed for both student and teacher inputs since our model will learn the discrimination among various noised views.
Notice that the mixed precision might lead to nanloss during training, so we only apply mixed precision for evaluation on small scale backbone (DiT-S) and transfer to full precision for large scale backbone (DiT-B and DiT-XL). All experiments are conducted on 8$\times$ 80GB-A100 GPUs.

\noindent\textbf{Evaluations.}\quad
To evaluate both the diversity and quality of our generative model, we utilize the most commonly adopted Fr\'echet Inception Distance (FID)~\citep{heusel2017gans} as evaluation metric.
For fair comparison with previous works~\cite{peebles2023scalable,zheng2023fast,gao2023masked}, we report FID-50K from ADM's TensorFlow evaluation suite~\cite{dhariwal2021diffusion} with the reference batch. We report the FID scores of the class-conditional sampling.  Besides, we provide more supporting metrics including Inception Score (IS)~\citep{salimans2016improved}, sFID~\citep{nash2021generating} and Precision/Recall~\citep{kynkaanniemi2019improved}.

\begin{figure}
    \centering
    \includegraphics[width=0.9\linewidth]{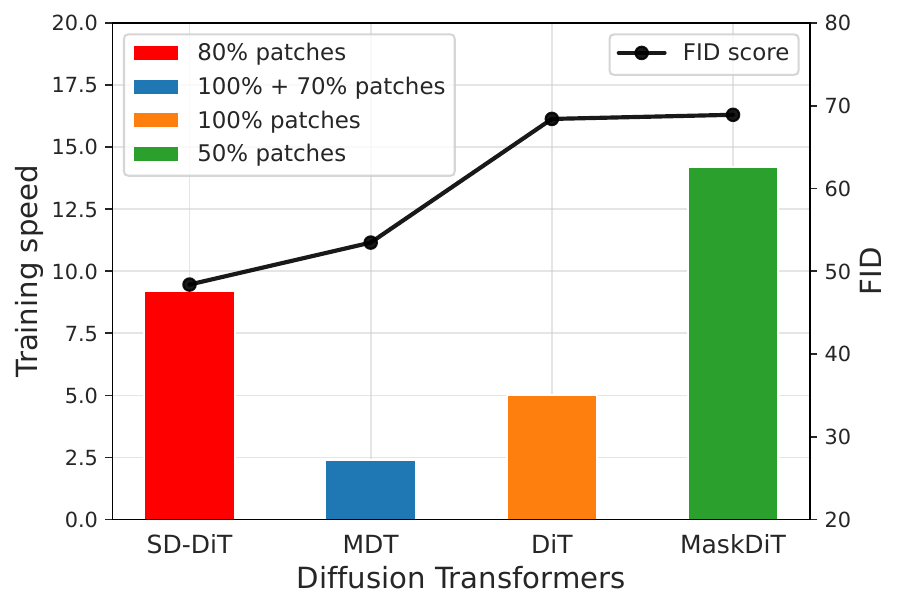}
    \vspace{-1.0em}
    \caption{Training speed (training steps per second) \emph{vs.} generative performance (FID-50K score) for our SD-DiT, MDT, DiT, and MaskDiT on 8 $\times$ A100 GPUs. We also label each run with the number of input patches.}
     \label{fig:speed}  
     \vspace{-2.0em}
\end{figure}

\subsection{Training Speed \emph{vs.} Performance}
Here we evaluate our SD-DiT with regard to both training speed and generative performance. \cref{fig:speed} shows the training speed (\ie, training steps per second) and FID-50K score of SD-DiT in comparison to state-of-the-art DiT models (DiT~\cite{peebles2023scalable}, MDT~\cite{gao2023masked}, and MaskDiT~\cite{zheng2023fast}) on 8 $\times$ A100 GPUs.
For fair comparison, the backbone network of each run is built on the same scale of DiT-S/2, same batch size (256) and training iterations (400k).
For SD-DiT and MaskDiT, we follow MDT and implement them with the same Float32 precision. For a comprehensive analysis, we also label each run with the number of input patches. 
As shown in \cref{fig:speed}, our SD-DiT (FID: 48.39; speed: 9.2 steps/sec or 0.11 sec/step) obtains better generative performance with faster training speed than MDT (FID: 53.46; speed: 2.4 steps/sec or 0.42 sec/step) and DiT (FID: 68.40, 5.03 steps/sec or 0.20 sec/step). 
This is due to that MDT simultaneously forwards and backwards both the complete (100\%) and visible patches (70\%), and DiT operates over the complete (100\%) patches, thereby resulting in slower training speed. 
In contrast, our SD-DiT and MaskDiT only forward and backward partial patches (80\%/50\%), leading to faster training speed. 
Furthermore, unlike MaskDiT that optimizes the whole encoder-decoder with mask reconstruction objective, our SD-DiT adopts a decoupled encoder-decoder structure to better exploit the mutual but also fuzzy relations between generative and discriminative objectives, leading to the best FID-50K score.
The results basically demonstrate the effectiveness of our SD-DiT which seeks a
competitive training speed-performance trade-off.

\begin{table}[t]
    \begin{minipage}[t]{1\linewidth}
        \small
        \centering
    \small
    \begin{tabular}{lccc}
    \toprule
       Method  &  Training Steps(k) & FID-50K$\downarrow$  \\
       \toprule
       DiT-S/2~\cite{peebles2023scalable}  & 400 & 68.40  \\
       MDT-S/2~\cite{gao2023masked}   & 400  &  53.46 \\
       \rowcolor{mygray} SD-DiT-S/2  & 400  &  \textbf{48.39}\\
       \midrule
       DiT-B/2~\cite{peebles2023scalable}  &  400 &  43.47 \\
       MDT-B/2~\cite{gao2023masked}  &  400 &  34.33 \\
       \rowcolor{mygray} SD-DiT-B/2  & 400  &  \textbf{28.62} \\
        \midrule
       DiT-XL/2~\cite{peebles2023scalable}  &  7000 &  9.62 \\
       MaskDiT-XL/2~\cite{zheng2023fast}  &  1300 &  12.15 \\
       MDT-XL/2~\cite{gao2023masked}  &  1300 &  9.60\\      
       \rowcolor{mygray} SD-DiT-XL/2  & 1100  & 9.66  \\
       \rowcolor{mygray} SD-DiT-XL/2  & 1300  & \textbf{9.01}  \\
       \bottomrule
    \end{tabular}
    \vspace{-0.8em}
    \caption{Performance comparison with state-of-the-art DiT-based approaches under various model sizes on ImageNet 256$\times$256 for class-conditional image generation (batch size: 256).}
    \label{tab:400k_results}
    \end{minipage}
    \hfill
    \begin{minipage}[t]{1\linewidth}
        \centering
        \includegraphics[width=0.85\linewidth]{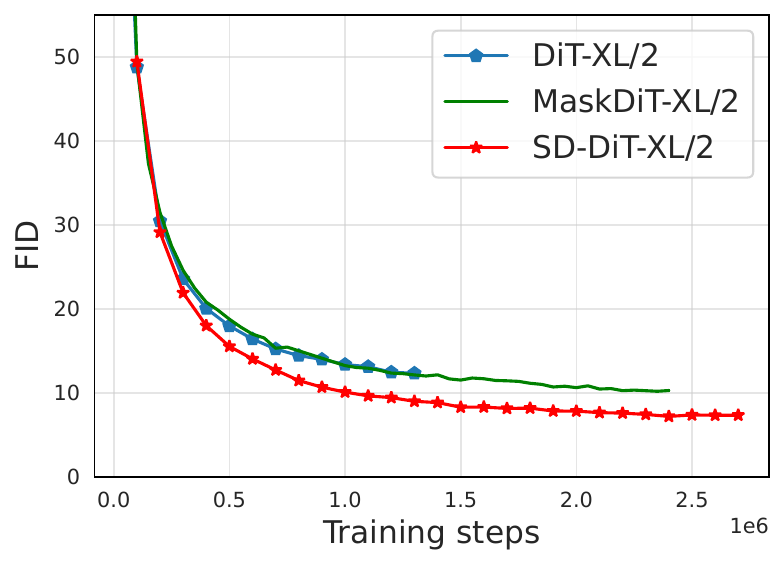}
        \vspace{-10pt}
        \captionof{figure}{Comparison of convergence speed with SOTA DiT-based approaches in DiT-XL backbone (batch size: 256). The results of DiT and MaskDiT are directly cited from MaskDiT~\cite{zheng2023fast}. Our SD-DiT-XL/2 consistently outperforms DiT-XL/2 and MaskDiT-XL/2 across training steps, leading to better training convergence.}
        \label{fig:across_step}   
     \centering
    \end{minipage}
    \vspace{-2.em}
\end{table}

\subsection{Performance Comparison} \label{sec:comparison}
\noindent\textbf{Comparison among Backbones in Different Scales.}\quad
\cref{tab:400k_results} provides comprehensive comparisons between our SD-DiT and several DiT-based state-of-the-arts under three different model sizes (DiT-S/B/XL). Notice that Mask-DiT only conducts experiments on DiT-XL backbone so we do not report its results on DiT-S and DiT-B backbones.
The batch size of all models is set as 256 for fair comparison.
Specifically, under the same small-scale backbone (DiT-S), our SD-DiT-S (48.39) exhibits better performance than DiT-S (68.40) and MDT-S (53.46) by a large margin.
This significant performance improvement of FID score is consistently observed when transferring to the larger scale backbones (DiT-B, DiT-XL). The results clearly validate the advantage of self-supervised discrimination knowledge distilling for mask modeling in Diffusion Transformer.

\noindent\textbf{Comparison on Convergence Speed in Large Scale Backbone.}\quad
Here we evaluate the convergence speed of our SD-DiT-XL/2 based on large-scale backbone. \cref{fig:across_step} illustrates the comparison of convergence speed by showing the FID scores in different training steps for our SD-DiT and various baselines. The batch size of each run is set as 256 for fair comparisons, and the maximum training step is 2400k. Note that the results of DiT and MaskDiT in different steps are directly copied from the reported results in MaskDiT~\cite{zheng2023fast}.
As shown in \cref{fig:across_step}, SD-DiT persistently reflects better training convergence than DiT and MaskDiT across the whole training steps. 
The detailed performance comparisons against MDT are listed in \cref{tab:400k_results}, where our SD-DiT (FID: 9.01) brings higher results than MDT (FID: 9.60) and MaskDiT (FID: 12.15) with 1300k training steps.
It is worthy noting that SD-DiT trained with 1300k steps outperforms typical DiT with 7000k steps (FID: 9.01 \emph{vs.} 9.62), achieving about 5$\times$ faster training progress.
In addition, SD-DiT (1100k steps) achieves a comparable FID performance with MDT (1300k steps) (9.66 \emph{vs.} 9.60). 
Such fast convergence again confirms the power of self-supervised discrimination for facilitating DiT training.

\noindent\textbf{Comparison with State-of-the-Art Generative Methods.}\quad
\cref{tab:sota} summarizes the performance comparison against state-of-the-art generative methods. 
We strictly follow MDT~\cite{gao2023masked} to list the cost comparison  column as ``Iter$\times$Batchsize''.
We follow the most DiT-based approaches and report the results in DiT-XL backbone with larger training iterations (2400k). 
Generally, under the same batch size of 256, our SD-DiT-XL/2 achieves a better FID score than DiT-XL/2 and MDT-XL/2. Although MaskDiT-XL/2 obtains the best FID score among all DiT-based methods, it benefits from the extremely large batch size of 1024. A more fair comparison between our SD-DiT and MaskDiT can be referred to \cref{fig:across_step}, where each run is trained with the same batch size (256). In that figure, SD-DiT-XL/2 leads to consistent performance boost against MaskDiT-XL/2, which clearly validates our proposal.

\begin{table}
    \centering
    \setlength{\tabcolsep}{0.19mm}  
    \small
    \begin{tabular}{lcccccc}
        \toprule
        Method	& Cost(Iter$\times$BS) &  FID$\downarrow$ & sFID$\downarrow$ & IS$\uparrow$ & Prec$\uparrow$ & Rec$\uparrow$ \\	\midrule
        VQGAN~\cite{esser2021taming} & - & 15.78 & 78.3 & - & - & - \\
        BigGAN-deep~\cite{brock2018large} & -  & 6.95 &  7.36 &  171.4 &  \textbf{0.87} &  0.28 \\
        StyleGAN~\cite{sauer2022stylegan} & -   & 2.30 & \textbf{4.02}  & \textbf{265.12} &  0.78 & 0.53 \\
        I-DDPM~\cite{nichol2021improved} & - & 12.26 & - & - & 0.70 & 0.62 \\
        MaskGIT~\cite{chang2022maskgit} & 1387k$\times$256 & 6.18 &  - & 182.1 & 0.80 & 0.51 \\
        CDM~\cite{ho2022cascaded}  & - & \textbf{4.88}  & - & 158.71 & - & - \\
        \midrule
        ADM~\cite{dhariwal2021diffusion}  &1980k$\times$256& 10.94 & 6.02 & 100.98 & 0.69 & 0.63 \\
        ADM-U~\cite{dhariwal2021diffusion}    &---    & 7.49 & 5.13 & 127.49 & 0.72 & 0.63 \\
        LDM-8~\cite{rombach2022high} & 4800k$\times$64 & 15.51 & - & 79.03 & 0.65 & 0.63 \\
        LDM-4~\cite{rombach2022high} & 178k$\times$1200 & 10.56  & -  & 103.49 & 0.71  & 0.62 \\
        \hline
        MaskDiT-XL/2\cite{zheng2023fast}    & 2000k$\times$1024   & 5.69  & 10.34    & 177.99      & 0.74    & 0.60  \\
        \hline
        DiT-XL/2~\cite{peebles2023scalable} & 7000k$\times$256  & 9.62  & 6.85  & 121.50  & 0.67  & \textbf{0.67} \\ 
        MDT-XL/2~\cite{gao2023masked} & 2500k$\times$256  & 7.41  & 4.95 & 121.22 & 0.72 & 0.64 \\
        \rowcolor{mygray} SD-DiT-XL/2 & 2400k$\times$256  &7.21  & 5.17 & 144.68 & 0.72 & 0.61 \\
        \bottomrule
    \end{tabular}
    \vspace{-0.8em}
    \caption{
    Performance comparison with state-of-the-art methods on ImageNet 256$\times$256 for class-conditional image generation. Similar to most DiT-based approaches, here we report the results of our SD-DiT in DiT-XL backbone with 256 batch size, while MaskDiT reports results with the largest batch size (1024).}
    \label{tab:sota}
    \vspace{-1em}
    \end{table}
\begin{table}[!t]
\small
	\centering
		\begin{tabular}{l|c}
			\toprule
                Method~&~~FID~~ \\ 
			\midrule
         	SD-DiT &  ~53.72  \\ 
			  w/o  Discriminative Objective ($\mathcal{L}_{\text{D}}$)~  & ~62.84\\ 
   			w/o Mask Strategy (mask ratio=0) & ~58.92\\ 
			\bottomrule
		\end{tabular}
        \vspace{-0.5em}
        \caption{Ablation studies on SD-DiT-S/2 with 400k training steps.}
         \vspace{-1.8em}
        \label{tab:ablation} 
\end{table}
\subsection{Ablation Study} \label{sec:ablation}
We conduct ablation study to examine each component in SD-DiT.
Considering that DiT training is computationally expensive, we adopt a lightweight setting for efficient evaluation: using small scale backbone (DiT-S) with 400k training steps, bs 256 and $50\%$ mask ratio unless specified.

\noindent\textbf{Effect of Discriminative Objective.}\quad
\cref{tab:ablation} details the performances of ablated runs of our SD-DiT. Specifically, the first row shows the FID score (53.7) of our complete SD-DiT-S/2 with $50\%$ mask ratio.
Next, by removing discriminative objective ($\mathcal{L}_{\text{D}}$) and the corresponding teacher branch from SD-DiT (2nd row), the generative performance drops by a large margin. 
This demonstrates the merit of our self-supervised discrimination tailored to Diffusion Transformer.
In addition, when removing mask strategy of SD-DiT (3rd row), a clear performance drop is attained, which highlights the effectiveness of mask strategy that triggers the learning of intra-image contextual awareness~\cite{zheng2023fast,gao2023masked}.

\begin{figure}[t]
    \centering
    \includegraphics[width=0.95\linewidth]{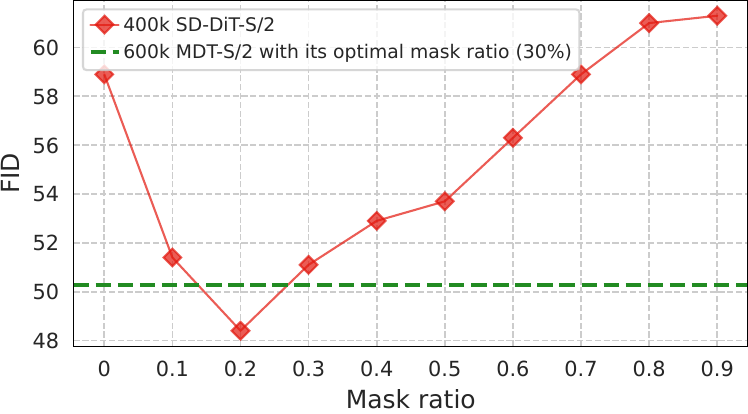}
    \vspace{-0.8em}
    \caption{FID \emph{vs.} mask ratio  on SD-DiT-S/2 with 400k steps.}
     \label{fig:mask}  
     \vspace{-1.8em}
\end{figure}

\noindent\textbf{Effect of Mask Ratio.}\quad
To further seek the sweet point of the balance between generative and discriminative task, we vary mask ratio from 0 to 1 and show the corresponding FID scores in \cref{fig:mask}. 
As shown in \cref{fig:mask}, the best performance of our SD-DiT is attained when the mask ratio is 20\%, and thus we adopt this ratio practically in all experiments of \cref{sec:comparison}.
We additionally show the performance of MDT-S/2 trained with 600K steps under its optimal 30\% mask ratio (the fixed green dashed line in \cref{fig:mask}, 50.3), which is inferior to our SD-DiT-S/2 with 400k steps (20\% mask ratio, 48.4). 
Moreover, MaskDiT points out one interesting observation with regard to mask ratio: MaskDiT with 75\% mask ratio achieves an extremely degraded FID score (121.16 of MaskDiT-XL/2). In other words, when 75\% patches participate in the mask reconstruction task and only 25\% local patches focus on the generative task, the generative ability of MaskDiT will be significantly weakened. This reveals the fuzzy relations between mask reconstruction and the generative task.
Instead, in our SD-DiT-S/2, even when the mask ratio is increased to 90\%, the corresponding FID score (61.0) is still higher than that of DiT-S/2 with 400k steps (68.40 in \cref{tab:400k_results}). These findings clearly verify that our design could alleviate the negative effect of fuzzy relations between mask modeling and generative task.

\begin{figure}[t]
    \centering
\includegraphics[width=0.95\linewidth]{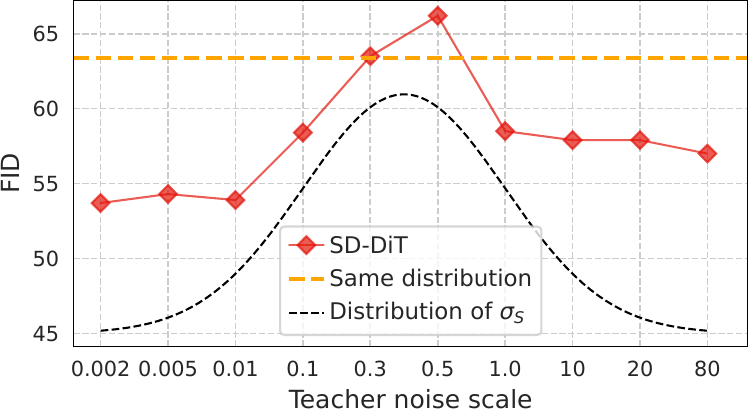}
    \vspace{-0.8em}
    \caption{FID \emph{vs.} teacher noise  on SD-DiT-S/2 with 400k steps.}
     \label{fig:teachernoise}  
      \vspace{-1.5em}
\end{figure}

\noindent\textbf{Effect of Noise of Teacher View.}\quad
Recall that in our SD-DiT, the noise of student view is set as $\x_{\sigma_\text{S}}=\x_0 + \n , ~\n \sim \mathcal{N}(\0,\sigma_\text{S}^2\I), ~\sigma_\text{S} \in[\sigma_\text{min},\sigma_\text{max}]$ based on EDM formulation (\cref{addnoise}).
Following EDM~\cite{karras2022elucidating} and Consistency Model~\cite{song2023consistency,song2023improved}, we set $\sigma_\text{min} = 0.002$ and $\sigma_\text{max} = 80$. Here we further test the effect of noise of teacher view. Specifically, we first set the noise of teacher view from the same distribution as student view, \ie,  $\x_{\sigma_\text{T}}=\x_0 + \n , ~\n \sim \mathcal{N}(\0,\sigma_\text{T}^2\I), ~\sigma_\text{T} \in[\sigma_\text{min},\sigma_\text{max}]$. As depicted in the yellow dashed line in \cref{fig:teachernoise}, the corresponding FID (63.4) is somewhat unsatisfying. This result shows that teacher noise derived from the same distribution of student noise can not make the discriminative loss practical for generative task. Such observation aligns with InfoMin principle~\cite{tian2020makes} in self-supervised learning:
reducing the mutual information between two variant views can bring a good pre-train model learning with sufficient view-invariance. That's why we choose the fixed minimum noise as in Consistency Models~\cite{song2023consistency,song2023improved} for teacher view, 
\ie, $\x_{\sigma_\text{T}}=\x_0 + \n , ~\n \sim \mathcal{N}(\0,\sigma_\text{min}^2\I)$.
In this way, the noise distribution of teacher view can be the closest one to the original data distribution ($\x_{\sigma_\text{T}} \sim  p_\text{data}(\x)$) and far away from student view.
We empirically evaluate various teacher noise within $[\sigma_\text{min},\sigma_\text{max}]$ (see the red curve in \cref{fig:teachernoise}), and the fixed minimum noise (scale: 0.002) can get the best performance (53.7).
Furthermore, we draw the approximate log-normal probability density distribution (PDF) of $\sigma_\text{S}$ based on EDM (see the black dashed line in \cref{fig:teachernoise}).
When the fixed $\sigma_\text{T}$ is set within the scale with high density (\eg, 0.3 and 0.5 close to the mean of $\sigma_\text{S}$), the corresponding FID of SD-DiT drops drastically (\eg, 66.2 when $\sigma_\text{T}=0.5$) and is even worse than the case of the same distribution (yellow dashed line).
This again reveals that the noise scale of teacher view should be far away from the distribution of $\sigma_\text{S}$.

\section{Conclusions}
In this work, we propose a Diffusion Transformer architecture, namely SD-DiT, to facilitate the training process by unleashing the power of self-supervised discrimination. SD-DiT novelly frames mask modeling in a teacher-student manner to jointly execute discriminative and generative diffusion processes in a decoupled encoder-decoder structure. Such design nicely explores the mutual but also fuzzy relations between mask modeling and generative objective, leading to both effective and efficient DiT training. Experiments conducted on ImageNet validate the competitiveness of SD-DiT when compared to SOTA DiT-based approaches.

\clearpage
{
    \small
    \bibliographystyle{ieeenat_fullname}
    \bibliography{main}
}
\clearpage

\begin{center}
		\LARGE
		\textbf{Appendix}
\end{center}
	
\maketitle
	
\appendix

\begin{figure}
    \centering
    \includegraphics[width=1\linewidth]{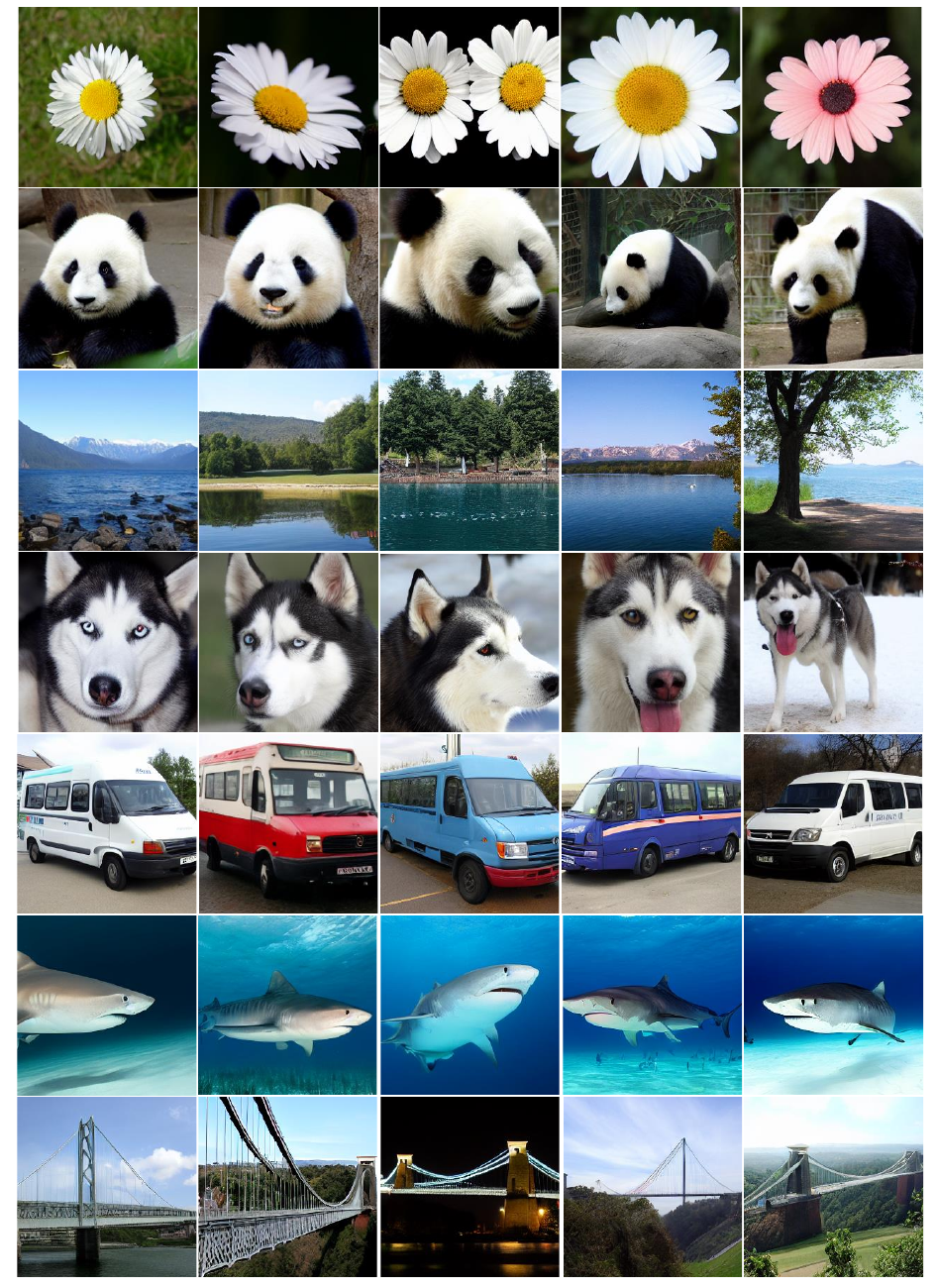}
    \caption{Qualitative results of our SD-DiT-XL/2. Label of each row (from top to bottom): Daisy, Giant panda, Lakeside, Eskimo dog, Minibus, Tiger shark, Suspension bridge.
 }
    \label{samples}
\end{figure}

The appendix contains: 1) more implementation details about our proposed Diffusion Transformer model with Self-supervised Discrimination (SD-DiT) in \cref{details}; 2) the pseudocode of SD-DiT in \cref{fage}; 3) the qualitative visualization results of SD-DiT-XL/2 in \cref{samples}.

\section{Implementation Details} \label{details}
In contrast to DiT~\cite{peebles2023scalable} and MDT~\cite{gao2023masked} whose settings are derived from the ADM formulation~\cite{dhariwal2021diffusion},
our SD-DiT employs the formulation of EDM~\cite{karras2022elucidating} in order to construct the discriminative pairs according to the theory of the \emph{consistency function} (Eq.(7) in main paper)~\cite{song2023consistency} based on the PF-ODE (Eq. (4) in main paper) of EDM.
Specifically, we adopt the EDM preconditioning parameterization by using a $\sigma$-dependent skip connection\footnote{Please refer to EDM~\cite{karras2022elucidating} for more comprehensive details.}:
\begin{equation}
\small
D_\theta(\x; \sigma) = \pmb{c}_\text{skip}(\sigma) ~\x + \pmb{c}_\text{out}(\sigma) ~F_\theta \big( \pmb{c}_\text{in}(\sigma) ~\x; ~\pmb{c}_\text{noise}(\sigma) \big).
\label{eq:preconditioning}
\end{equation}
This preconditioning parameterization is a common practice to 
avoid large variation in gradient magnitudes brought by various noise levels.
As shown in \cref{eq:preconditioning}, the denoiser $D_\theta$ is not directly employed as a neural network. Instead, a different network $F_\theta$ is trained to learn $D_\theta$.
In our SD-DiT, the student branch is wrapped as $D_\theta$
in \cref{eq:preconditioning} with skip connection preconditioning.
For simplicity, we did not introduce this parameterization in the main paper.
We follow the default hyper-parameters of EDM for the skip connection $\pmb{c}_\text{skip}(\sigma)$, the noise level $\pmb{c}_\text{noise}(\sigma)$ and the input $\pmb{c}_\text{in}(\sigma)$ and output magnitudes   $\pmb{c}_\text{out}(\sigma)$.
Besides, the student noise distribution $p_{\sigma_\text{S}}$
follows the $p_{\sigma_\text{train}}$ in EDM's setting:
\begin{equation}
\label{ptrain}
    \ln(p_{\sigma_\text{S}})\sim \mathcal{N}(P_\text{mean},P_\text{std}), 
\end{equation}
where $P_\text{mean}=-1.2$ and $P_\text{std}=1.2$. 
Note that we draw the approximate log-normal probability density distribution (i.e., the black dashed line in Fig. 5 in main paper) of the corresponding $\sigma_\text{S}$ according to this \cref{ptrain}. 
During the sampling stage, 
we use the default time steps schedule of EDM: 
\begin{equation}
\label{eq:discretization}
\small
\sigma_{i<N} = \big( {\sigma_\text{max}}^\frac{1}{\rho} + {\textstyle\frac{i}{N-1}} ( {\sigma_\text{min}}^\frac{1}{\rho} - {\sigma_\text{max}}^\frac{1}{\rho} ) \big)^\rho, \sigma_N = 0 \text{,}
\end{equation}
where sampling steps $N=40$, $\rho=7$, $\sigma_\text{min}=0.002$ and $\sigma_\text{max}=80$. 
Following EDM, we utilize the second-order Heun ODE solver for sampling.
We follow the paradigm of LDM~\cite{rombach2022high} to perform diffusion generation in the latent space of the frozen pre-trained VAE model~\cite{rombach2022high}, which downsamples a $256\times256\times3$ image into a $32\times32\times4$ latent variable.  
More implementation details can be referred in \cref{tab:hyperparams}.

\noindent\textbf{Network parameters.}\quad
The teacher-student design will double the parameters of a typical DiT. But at inference,  the teacher network will be removed, and thus no parameter burden is introduced. In this sense, the model size of learned SD-DiT-XL/2 is 740.6M, which is comparable to MaskDiT-XL/2 (730.1M). 
During training, the additional teacher network is directly updated by EMA without SGD backward propagation, thereby only requiring extremely lightweight computational cost compared to standard backward propagation. At inference, the teacher network is completely removed and no burden is introduced.

\begin{table}[H]
\centering
\small
\caption{
Configs for training SD-DiT on 256$\times$256 ImageNet-1K.
}
\vspace{-1em}
\resizebox{0.49\textwidth}{!}{
\begin{tabular}{l|ccc}
\toprule
\bf Configs & SD-DiT-S/2  & SD-DiT-B/2 & SD-DiT-XL/2   \\
\midrule
total batch size & \multicolumn{3}{c}{256} \\
learning rate & \multicolumn{3}{c}{1e-4} \\
training iterations  & 400k & 400k & 2400k \\
optimizer & \multicolumn{3}{c}{AdamW~\cite{loshchilov2019decoupled} with $\beta_1, \beta_2{=}0.9, 0.999$} \\
EMA momentum & \multicolumn{3}{c}{from 0.996 to 0.999} \\
student temperature & \multicolumn{3}{c}{0.1} \\
teacher temperature & \multicolumn{3}{c}{from 0.09 to 0.099 (warmup 5 epochs)} \\
\bottomrule
\end{tabular}}
\vspace{-1em}
\label{tab:hyperparams}
\end{table}

\section{Additional Experimental Results} \label{other_exp}
\paragraph{How about training with a larger batch size?}\quad
MaskDiT-XL/2 attains the best FID score with fewer training steps, attributed to a large batch size of 1024.
For a more comprehensive comparison,
we experiment by training SD-DiT-XL/2 with 1024 batch size, and the FID is 16.78 (150k steps), which is better than MaskDiT-XL/2 (FID: 17.22 at 150k steps)~\cite{zheng2023fast}. 

\paragraph{Comparison at higher iterations.}\quad
We experiment by training SD-DiT-XL/2 with higher iterations (3500k), and the FID is 6.74, which is comparable to MDT-XL/2 (FID: 6.65, 3500k)~\cite{gao2023masked}.
It is worth noting that, compared to , our SD-DiT-XL/2 only uses 45GB memory per GPU with faster training speed (much lower than the memory requirement of MDT-XL/2~\cite{zheng2023fast}), leading to a better computational cost-performance trade-off. 

\paragraph{Classifier-free guidance (CFG) results.}\quad
We also experiment by upgrading our SD-DiT with CFG, and the FID of SD-DiT-XL/2 (+CFG) is 3.23, which is better than MaskDiT-XL/2 (without the unmask tuning stage) with CFG (FID: 4.54)~\cite{zheng2023fast}.

\section{Pytorch-like Pseudocode for SD-DiT} \label{pcodes}
\begin{lrbox}{\mycode}
\begin{lstlisting}[language=Python, escapeinside={(*}{*)}]
# (*$\mathcal{S}_\theta$*),(*$\mathcal{T}_\theta'$*): student & teacher DiT encoder 
# (*$\mathcal{G}_\theta$*): student DiT decoder 
# (*$j_{\theta}$*), (*$j_{\theta'}$*): student and teacher MLP projection head
# C_cls: center (K) for cls dicrimitive loss
# C_patch: center (K) for patch dicrimitive loss
# (*$\tau_\text{S}$*), (*$\tau_\text{T}$*): student and teacher temperatures
# (*$\beta$*), m_c, m_p: the momentum rates of network, center of C_cls, and center of C_patch   
# (*$\x_0$*): We extract all the latents from raw image by VAE encoder to directly model our SD-DiT on the basis of Latent Diffusion Model.

(*$\mathcal{T}_\theta'$*).params = (*$\mathcal{S}_\theta$*).params
for (*$\x_0$*) in loader: # load a minibatch with N samples
    
    # construct noised student and teacher views
    (*$\x_{\sigma_\text{S}}=\x_0 + \n_\text{S}, ~\n_\text{S} \sim \mathcal{N}(\0,\sigma_\text{S}^2\I), ~\sigma_\text{S} \in[\sigma_\text{min},\sigma_\text{max}]$*) 
    (*$\x_{\sigma_\text{T}}=\x_0 + \n_\text{T} , ~\n_\text{T} \sim \mathcal{N}(\0,\sigma_\text{min}^2\I)$*)
    # random mask (*$\mathcal{M}$*) for student view
    (*$\vv_{\sigma_\text{S}}=\x_{\sigma_\text{S}}\odot(1-\mathcal{M})$*) # visible patches 
    (*$\pmb{\bar v}_{\sigma_\text{S}} =\x_{\sigma_\text{S}}\odot\mathcal{M}$*)      # invisible patches 
    # forward student and tecaher encoder
    (*$\e_\text{S}=\mathcal{S}_{\theta}(\vv_{\sigma_\text{S}})$*)  # forward visible student patches 
    (*$\e_\text{T}=\mathcal{T_{\theta'}}(\x_{\sigma_\text{T}})$*)  # forward full teacher patches 

    ######## Generative Loss ########
    # insert invisible patches onto visible tokens according to mask positions
    (*$\mathcal{H}=$*) torch.gather(torch.cat((*$\e_\text{S}$*),(*$\pmb{\bar v}$*)), (*$\mathcal{M}$*))
    # feed complete token set to student decoder
    (*$\pmb{o_\text{S}}=\mathcal{G}_\theta(\mathcal{H}) $*) # output all tokens for generative loss
    (*$\mathcal{L}_{\text{G}}=$*) MSELoss((*$\pmb{o_\text{S}}$*), (*$\x_0$*)).mean()
    
    ######## Discriminative Loss ########
    # forward projection head
    (*$j(\e_\text{S}^{\text{[cls]}})$*), (*$j(\e_\text{S}^{\text{patch}})$*) = (*$j_{\theta}$*)((*$\e_\text{S}$*)) #cls token dim: [N,1,K]
    (*$j(\e_\text{T}^{\text{[cls]}})$*), (*$j(\e_\text{T}^{\text{patch}})$*) = (*$j_{\theta'}$*)((*$\e_T$*)) #patch token dim:[N,L,K]
    # inter-view discriminative loss on CLS token 
    (*$\mathcal{L}_{\text{D}}^{\text{cls}}$*)= H((*$j(\e_\text{T}^{\text{[cls]}})$*), (*$j(\e_\text{S}^{\text{[cls]}})$*), C_cls)
    # inter-view discriminative loss on patch tokens 
    (*$\mathcal{L}_{\text{D}}^{\text{patch}}$*)= H((*$j(\e_\text{T}^{\text{patch}})$*), (*$j(\e_\text{S}^{\text{patch}})$*), C_patch) 
    
    ##############################################
    Loss = (*$\mathcal{L}_{\text{G}}$*) + (*$\mathcal{L}_{\text{D}}^{\text{cls}}$*) + (*$\mathcal{L}_{\text{D}}^{\text{cls}}$*) 
    Loss.backward() # back-propagate
    update((*$\theta$*)) # SGD update for student branch
    
    # teacher and center updates
    (*$\theta'$*).params = (*$\beta$*)*(*$\theta'$*).params + (1-(*$\beta$*))*(*$\theta$*).params
    # center updates by teacher patches and cls token 
    C_cls = m_c*C_cls + (1-m_c)*(*$j(\e_\text{T}^{\text{[cls]}})$*).mean(dim=0)
    C_patch = m_p*C_patch + (1-m_p)*(*$j(\e_\text{T}^{\text{patch}})$*).mean(dim=0,1)

def H(T, S, C): # cross-entropy loss
    T = T.detach() # stop gradient
    S = softmax(S/(*$\tau_\text{S}$*), dim=1)
    T = softmax((T - C)/(*$\tau_\text{T}$*), dim=1) # center + sharpen
    return - (T * log(S)).sum(dim=1).mean()



\end{lstlisting}
\end{lrbox}

\begin{algorithm}[t]
\usebox{\mycode}
\caption{Pytorch-like Pseudocode of SD-DiT}
 \label{fage}
 \algcomment{
\textbf{Notes}: 
Note that patch-level discriminative loss is solely performed over the visible patch tokens.
Here we do not show them in the pseudocode for simplicity. Moreover, we do not show the skip-connection preconditioning in this pseudocode.
}
\end{algorithm}

\end{document}